\documentclass{article}

\usepackage[final, nonatbib]{nips_2018}

\usepackage[utf8]{inputenc} 
\usepackage[T1]{fontenc}    
\usepackage{hyperref}       
\usepackage{url}            
\usepackage{booktabs}       
\usepackage{amsfonts}       
\usepackage{nicefrac} 
\usepackage{microtype}    
\usepackage{subfigure}
\usepackage{multirow}
\usepackage{tikz}
\usepackage{subfiles}
\usepackage{graphicx}
\usepackage{xcolor}
\usepackage{amsmath}
\usepackage{amsfonts}
\usepackage{sansmath}
\usepackage[export]{adjustbox}

\definecolor{blue_1}{RGB}{0,68,129}
\definecolor{blue_4}{RGB}{91,190,255}
\definecolor{red_1}{RGB}{185,42,69}
\definecolor{red_4}{RGB}{231,125,142}

\title{Towards Global Explanations for Credit Risk Scoring}

\author{
  Irene Unceta \\
  BBVA Data \& Analytics \\
  \texttt{irene.unceta@bbvadata.com} \\
  \And
  Jordi Nin \\
  BBVA Data \& Analytics \\
  \texttt{jordi.nin@bbvadata.com} \\
  \And
  Oriol Pujol\\ 
  Universitat de Barcelona \\
  \texttt{oriol\_pujol@ub.edu} \\
}

\begin{document}

\maketitle

\begin{abstract}
In this paper we propose a method to obtain global explanations for trained black-box classifiers by sampling their decision function to learn alternative interpretable models. The envisaged approach provides a unified solution to approximate non-linear decision boundaries with simpler classifiers while retaining the original classification accuracy. We use a private residential mortgage default dataset as a use case to illustrate the feasibility of this approach to ensure the decomposability of attributes during pre-processing.
\end{abstract}

\section{Introduction}
\label{sec:intro}

Failure to keep up with loan repayments, also known as credit default, has significant cost implications for financial institutions, which devote increasing efforts to develop and refine credit scoring models to predict defaults~\cite{PRESSMonth}. Such efforts result in an non-trivial models, which are able learn more complex problems with a high degree of accuracy. However, automated decision making by banks is subject to great scrutiny, with financial regulators~\cite{Goodman2017EuropeanExplanation} requiring, among other specifications, that internal coefficients and variable importance be accessible and in line with human domain knowledge~\cite{FSB2017FinancialAttention}. This is why a widely established technique for credit risk modelling is Logistic Regression (LR). Indeed, models based on LR, while performing relatively well on default prediction settings, offer the additional advantage of a relative ease of interpretation. Not in vain do institutions such as FICO use LR for their risk scores, concretely mentioning interpretability as one of the main reasons for it~\cite{FairIsaacCorporationFICO2011IntroductionBuilder}. 

Even so, LR models are linear and fail to account for non-linearities in the data, at the cost of accuracy. To overcome this limitation, non-linear effects are usually modelled during pre-processing, when domain knowledge by experts is exploited to obtain a set of highly predictive artificially generated attributes. This practice is against the idea of \emph{intelligibility} as described in~\cite{Lou2012IntelligibleRegression} and results in \emph{non-decomposable}~\cite{Lipton2016TheInterpretability} machine learning pipelines. Previous works devoted to explaining how machine learning models work have attempted to generate local explanations around a set of predictions~\cite{Ribeiro2016WhyClassifier,Ribeiro2018Anchors:Explanations}. Evenwhile such explanations may faithfully represent the functioning of the model locally, they fail to meet regulatory requirements in that they are based on non-interpretable data attributes. The problem of balancing accuracy and interpretability in credit risk scoring therefore remains unsolved.

In this work, we introduce a novel approach to obtain global explanations based on copies that mimic the decision behavior of trained machine learning classifiers. We derive copies based on interpretable models that act in accordance with the compliance requirements. We argue that the process of copying differs from that of learning, as traditionally understood by the machine learning community. 

Our main contribution is a two-fold solution to the problem of interpreting high performing credit risk scoring models in BBVA. On the one hand, we erase the pre-processing step by deploying more complex algorithms that capture non-linear data relations. We deliver such models to production by substituting them with more flexible copies that remain interpretable, while retaining a good overall accuracy. On the other hand, we keep decomposability by building copies that learn the decision outputs of trained models directly from the raw attributes.

\section{Model-agnostic copies of trained machine learning classifiers}
\label{sec:methods}

In this section we describe the framework for copying machine learning classifiers. We use the term \textit{original dataset} to refer to a set of pairs $\boldsymbol{X} = \{(\boldsymbol{x}_i, t_i)\}, i = 1, ..., M$, where $x_i \in \mathbb{R}^d$ is a set of  $d$-dimensional data points in the \textit{original feature space} $\mathcal{D}$ and  $t_i \in \{1,\dots, K\}$ their corresponding labels. We define the \textit{original model}, $f_{\mathcal{O}}: \boldsymbol{X} \rightarrow \mathbf{t}$, as any model trained using $\boldsymbol{X}$. 

Our aim when copying is to reproduce the behavior of the original model by means of a {\it copy},  $f_{\mathcal{C}}$, such that  $f_{\mathcal{C}}(x) = f_{\mathcal{O}}(x),\; \forall x \in \mathcal{D}$. To do so we use a two step process. First, we rely in the generation of a {\it synthetic dataset}, $\boldsymbol{Z} = \{(\boldsymbol{z}_j, y_j)\}, j = 1, ..., N$, where $\boldsymbol{z}_j$ is sampled randomly from a probability distribution over the original feature space $\mathcal{D}$ and labelled according to the predictions of the original classifier, \textit{i.e.} $y_j = f_{\mathcal{O}}(\boldsymbol{z}_j)$. Second, we use the synthetic data points to train a copy, $f_\mathcal{C}$, whose decision function reproduces $f_{\mathcal{O}}$ to the extent that it can be used to substitute it.

\subsection{Copying is not learning}

The use of surrogate models is not new, specially when it comes to extracting rules from black-box methods~\cite{Bologna2018ASVMs,Craven1995ExtractingNetworks}. However, literature has traditionally treated extraction as a standard learning process, where original training data is relabelled and extended to learn an alternative model. In what follows, we argue that there exist fundamental differences between this approach and ours.

Following previous literature suggestions one might be tempted to use standard learning methods to copy a relabelled version of the original dataset. In our proposal, however, one is not necessarily tied to the original data. During the copying process, the original model translates the original problem into a hard classification boundary. Thus, separability of the synthetic dataset is always guaranteed. If we select a model with enough representation capacity, then it is possible to achieve zero empirical error, so that the error of the copy only depends on the generalization gap for the synthetic dataset.

Because in our framework we control the synthetic data generation process, we have access to a potentially infinite stream of synthetic points. As a result, the generalization gap can be asymptotically reduced to zero and copying can be performed without loss of accuracy. Indeed, we cast copying as an unconstrained optimization problem where we minimize the empirical risk $R_{emp}(h(\theta), \boldsymbol{Z})$ (or a surrogate loss function) in terms of model parameter $\theta$. In this context, when ensuring a good generalization of the copy, deploying typical practices to avoid overfitting, such as regularization, may negatively affect copy performance since they can easily lead to a reduced model capacity, effectively preventing the copy algorithm from reaching the minimum for $R_{emp}(h(\theta), \boldsymbol{Z})$. Moreover, hyperparameter optimization, is not needed. We argue that because of these fundamental differences, the concept of generalization has to be redefined in the copying scenario.

\subsection{Towards interpretable copies}

In every credit risk model there exists a complex trade-off between prediction accuracy and explainability: the former is of paramount importance, the latter a mandatory requirement. Therefore, a financial institution is hardly ever willing to give up one for the sake of the other. Here we propose the use of copies as a solution to this compromise. 

When copying, we can transfer model features from one model to another. Thus, we can ensure a high accuracy is retained by exploiting the predictive performance of high capacity models during training of the original and then augment this capacity by substituting the original with a self-explanatory copy. Provided we are able to faithfully mimic the original decision function, the copying process can be interpreted as endowing the original model with features of the copy. This effectively bridges the trade-off between accuracy and interpretability by generating a model that displays both features. 

\begin{figure}
    \centering
    \includegraphics[width=5.5in]{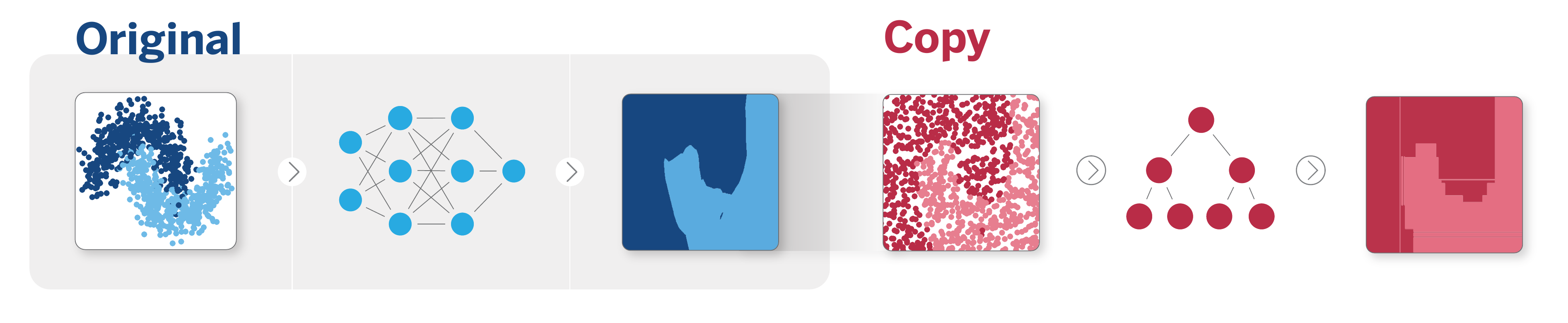}
    \caption{A binary classification dataset is fed to a neural network, whose predictions are used to label synthetic data points sampled uniformly at random from the original feature space. These points are then used to train a decision tree whose decision behavior resembles that of the original network.}
\label{example}
\end{figure}

An example of this is shown in \emph{Figure \ref{example}}, where an original dataset is learned by a neural network that yields a classification accuracy of 0.96 and copied using a decision tree that obtains an accuracy of 0.93 over the same test data. The complexity of the considered problem is captured by the more complex structure of the original network, which is trained to yield an optimal performance. This structure is then substituted by the simpler copy model, that has the added advantage of being interpretable. \\

\section{Use cases}
\label{sec:experiments}

In this section we validate our proposal in the context of non-client residential mortgage loan default prediction. We use a private dataset consisting of the 19 financial attributes described in the supplementary material for 1.328 non-client loan applications recorded by BBVA during 2015 all over Mexico. Though at the time of loan application all individuals in this dataset were considered by the bank to be creditworthy and therefore granted a mortgage loan, only 1025 of them paid it off, which corresponds to a ratio of defaulted loans equal to 0.23. 

Due to the sensitive nature of bank data, we anonymize and identify all customers using randomly generated IDs. Additionally, we convert all nominal attributes to numerical and re-scale all features to zero-mean and unit variance. We perform a stratified split to obtain training and test sets with relative sizes of 0.8 and 0.2, respectively. Synthetic datasets are generated drawing random samples from a uniform distribution defined on the original feature space $\mathcal{D}$ and labelling them according to the predictions output by an original classifier. In all cases, we use training and test synthetic sets comprised of $N=10^6$ samples. We train copies based on decision trees (DT) and report results averaged over 100 different runs. Following the discussion in \textit{Section \ref{sec:methods}}, we enforce no capacity control on the copy algorithms.

We suggest two different approaches to deliver interpretable machine learning solutions that yield a good prediction performance for this problem, while complying with regulatory requirements. In \textit{Scenario 1}, we use copies to ensure the attributes of a risk scoring model remain intelligible. In \textit{Scenario 2}, we avoid the pre-processing step by exploiting a higher learning capacity model and then copying it with a simpler yet interpretable one.

\subsubsection*{Scenario 1: Keeping decomposability on mortgage loan default models}

We emulate a standard risk modelling production pipeline by pre-processing the original dataset to obtain 4 new variables. We use these variables together with \textit{age} and \textit{economy\_level} to learn a LR model. Note that in a real setting, to obtain the final dataset used to train this LR, a tedious process of trial and error is conducted by a qualified risk analyst until a valid set of predictive variables is found. This incurs in a large economical cost and a delayed time-to-market delivery. Even worse, this pre-processing largely reduces the interpretability of the learned LR model, since the new variables often reflect complex relations among the original data attributes.

The whole predictive system, composed by both the pre-processing/feature engineering step and the LR model, yields an accuracy of $0.77$ for original test data. We copy this system by means of a DT with a mean accuracy of $0.71 \pm 0.04$ on original test data and the accuracy distribution shown \textit{Figure \ref{features} (b)}. Thus, we substitute the original pipeline with a non-linear yet interpretable function that is directly applied on the raw input features and which outputs the learned prediction labels. As a result of this process, we can provide new explanations based on the original data attributes and thus, maintain the decomposability of both the feature crafting process and the original machine learning technique used.

\subsubsection*{Scenario 2: High performance self-explanatory risk scoring} 
 
\begin{figure}
\centering
\begin{tabular}{cc}
  \multirow{2}{*}{\adjustbox{valign=t}{\includegraphics[width=3.8in, trim={1.5cm 0.5cm 1.5cm 0.5cm}]{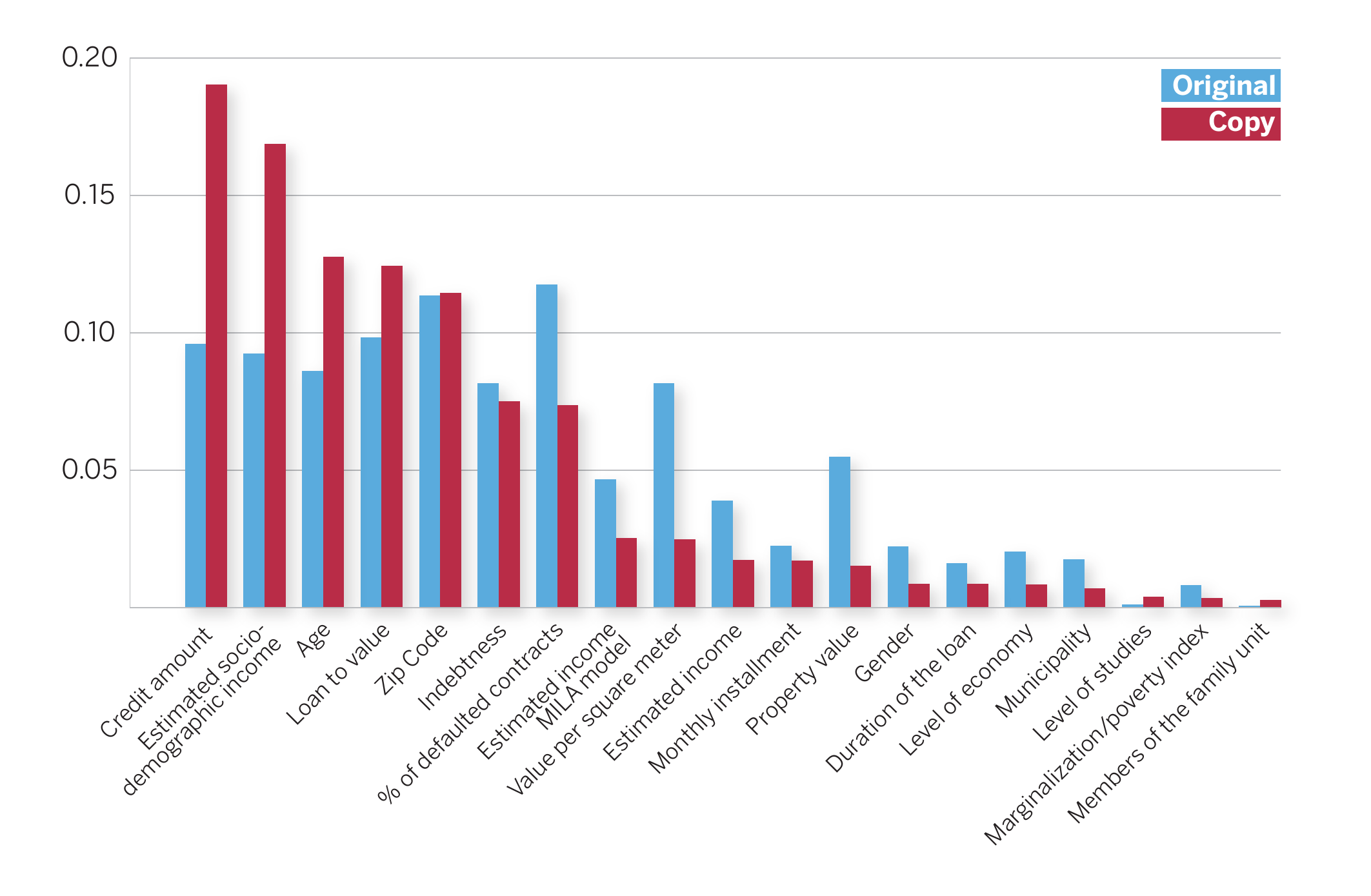}}} & \adjustbox{valign=t}{\includegraphics[width=1.2in, trim={1cm 0.5cm 1cm 0.5cm}]{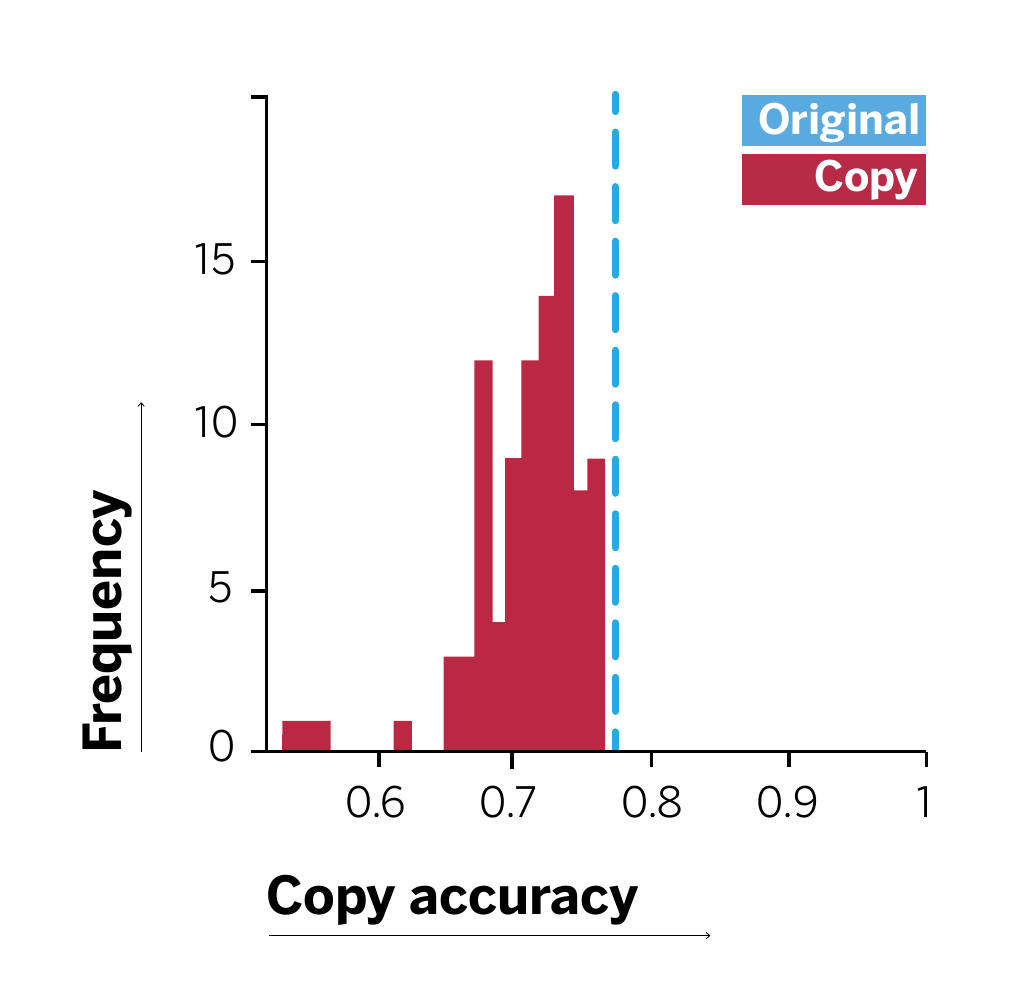}} \\
  & (b) \\
  & \includegraphics[width=1.2in, trim={1cm 0.5cm 1cm 0.25cm}]{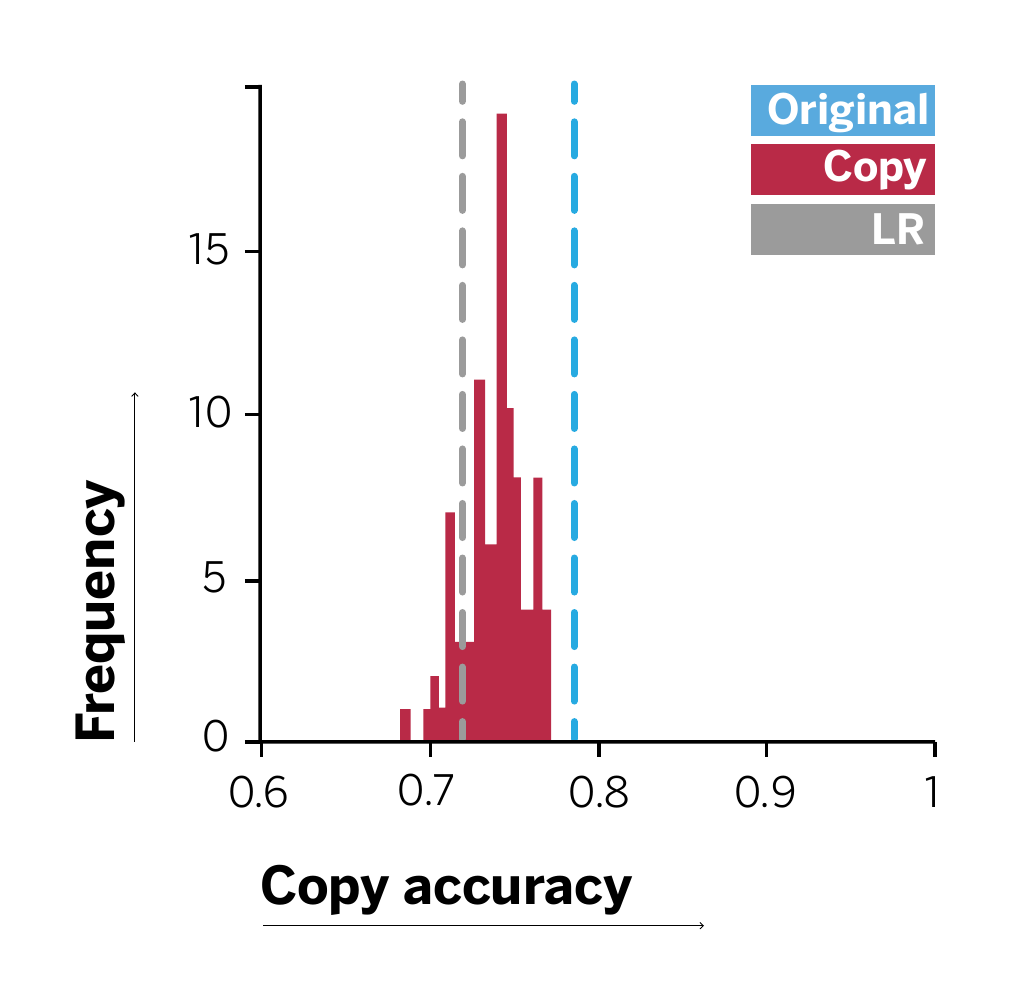} \\
  (a) & (c) \\
\end{tabular}
\caption{(a) Feature importance for the original gradient-boosted tree (blue), the copy decision tree (orange) and superposition of both. Distribution of copy accuracies for scenarios (b) 1 and (c) 2.}
\label{features}
\end{figure}

We use all the 19 features depicted in the supplementary material to train a gradient-boosted tree that yields an accuracy of $0.79$. This value is sensibly higher than that obtained by the pre-processed LR above. This is because the learned decision function is able to capture non-linear relationships among original data attributes. However, because of the boosting step, the resulting model is not easily interpretable. We copy it using a standard DT that yields a mean accuracy of $0.739 \pm 0.018$ for original test data, corresponding to the distribution displayed in \textit{Figure \ref{features} (c)} . When LR is applied directly to raw attributes, it achieves an accuracy equal to 0.72. 

In \textit{Figure \ref{features} (a)} we show the feature importance for both original and copy models, ranked in terms of the latter. Even while punctual differences can be observed in the importance scores of certain variables, both models agree on which the most important attributes are. Notably, the distribution of importance scores is more skewed for the copy, meaning that after the copying process, errors are minimized from a global perspective and it is possible to better discriminate among features. 

\section{Conclusions}
\label{sec:conclusions}

In this paper we show how model-agnostic copies of machine learning classifiers can be used to tackle the problem of interpretability in credit risk scoring. Copies endow new features not present in original models while retaining their performance. We briefly sketch the reasons why copying departs from standard machine learning assumptions and practices and argue that the success of this process relies on two concepts. First, we emphasize the crucial role of the synthetic dataset to capture the global behavior of the original model. Secondly, we discuss the particularities of the optimization process for building a copy. Finally, we validate our proposal in two real scenarios: decomposability of the feature domain engineering and regulatory compliant high performance model building. 

We conclude this paper by identifying open issues that deserve further attention. There exist at least two sources of loss in the copying process that arise from the synthetic sample generation itself. For one, although having access to an infinite stream of data, in practice we are limited to a large but finite set of synthetic data points. Also, ensuring an optimal exploration of the original feature space becomes challenging for increasing problem dimensionality. We would like to extend this work to derive efficient algorithms to generate optimal synthetic datasets, as well as to study the exact performance guarantees of the copying process in terms of the different sources of error.

\subsubsection*{Acknowledgments}

This work has been partially funded by the Spanish project TIN2016-74946-P (MINECO/FEDER, UE), and by AGAUR of the Generalitat de Catalunya through the Industrial PhD grant 2017-DI-25. We gratefully acknowledge the support of BBVA Data \& Analytics for sponsoring the Industrial PhD.

\bibliographystyle{plain}
\bibliography{main.bbl}

\end{document}